\documentclass[conference]{IEEEtran}
\IEEEoverridecommandlockouts
\usepackage{cite}
\usepackage{amsmath,amssymb,amsfonts}
\usepackage{graphicx}
\usepackage{textcomp}
\usepackage{xcolor}
\usepackage{multirow}
\usepackage{arydshln}
\usepackage{todonotes}
\usepackage{etoolbox}
\usepackage{setspace}
\usepackage[bottom]{footmisc}
\usepackage{placeins}
\usepackage{wrapfig}
\usepackage{color, colortbl}
\usepackage[export]{adjustbox}
\usepackage{amsmath}
\usepackage{algpseudocode}
\usepackage[ruled,vlined]{algorithm2e}
\usepackage{tikz}
\usepackage{relsize}
\usepackage{mathtools, nccmath}
\def\BibTeX{{\rm B\kern-.05em{\sc i\kern-.025em b}\kern-.08em
    T\kern-.1667em\lower.7ex\hbox{E}\kern-.125emX}}
    
\usepackage{url}

\begin{document}
    \title{
   Efficient and Compact Convolutional Neural Network Architectures for Non-temporal Real-time Fire Detection
}
\author{\IEEEauthorblockN{William Thomson$^{1}$,
   Neelanjan Bhowmik$^{1}$,
   Toby P. Breckon$^{1,2}$
   }
 \IEEEauthorblockA{Department of \{Computer Science$^1$ $|$ Engineering$^2$\}, Durham University, UK}
 }

\maketitle

\begin{abstract}
Automatic visual fire detection is used to complement traditional fire detection sensor systems (smoke/heat). In this work, we investigate different Convolutional Neural Network (CNN) architectures and their variants for the non-temporal real-time bounds detection of fire pixel regions in video (or still) imagery. Two reduced complexity compact CNN architectures (NasNet-A-OnFire and ShuffleNetV2-OnFire) are proposed through experimental analysis to optimise the computational efficiency for this task. The results improve upon the current state-of-the-art solution for fire detection, achieving an accuracy of 95\% for full-frame binary classification and 97\% for superpixel localisation. We notably achieve a classification speed up by a factor of 2.3$\times$ for binary classification and 1.3$\times$ for superpixel localisation, with runtime of 40 fps and 18 fps respectively, outperforming prior work in the field presenting an efficient, robust and real-time solution for fire region detection. Subsequent implementation on low-powered devices (Nvidia Xavier-NX, achieving 49 fps for full-frame classification via ShuffleNetV2-OnFire) demonstrates our architectures are suitable for various real-world deployment applications.
\end{abstract}

    \section{Introduction} \label{sec:intro}
Automatic real-time fire (or flame) detection by analysing video sequences is increasingly deployed in a wide range of auto-monitoring tasks. The monitoring of urban and industrial areas and public places using security-driven CCTV video systems has given rise to the consideration of these systems as sources of initial fire detection (in addition to heat/smoke based systems). Furthermore, the on-going consideration of remote vehicles for automatic fire detection and monitoring tasks \cite{bardshaw91robots,MARTINEZDEDIOS2006uavfire} further enhances the demand for autonomous fire detection from such platforms. Detecting fire in images and video is a challenging task among other object classification tasks due to the inconsistency in its shape or pattern and varies with the underlying material composition.

Many earlier traditional approaches in this area involve attribute based approaches, such as the colour \cite{Healey93fire,CELIK2009fire} and they can be combined in high-order temporal approaches \cite{Phillips02fire,liu2004fire,Toreyin2005fire}. The work of \cite{Healey93fire} uses a colour based threshold approach on input video. This is expanded in the work \cite{Phillips02fire} which incorporates both colour and motion, utilising a colour histogram to classify fire pixels and examine the temporal variation of the pixels to determine which are fire. The work by \cite{liu2004fire} further explores temporal variation in the Fourier coefficients of fire regions to capture the contour of the region. Slightly more recent attribute based approach \cite{Toreyin2005fire} models flame coloured objects using Markov models to help distinguish the flames. 

\begin{figure}[htb!]
    \centering
    \includegraphics[width=\linewidth]{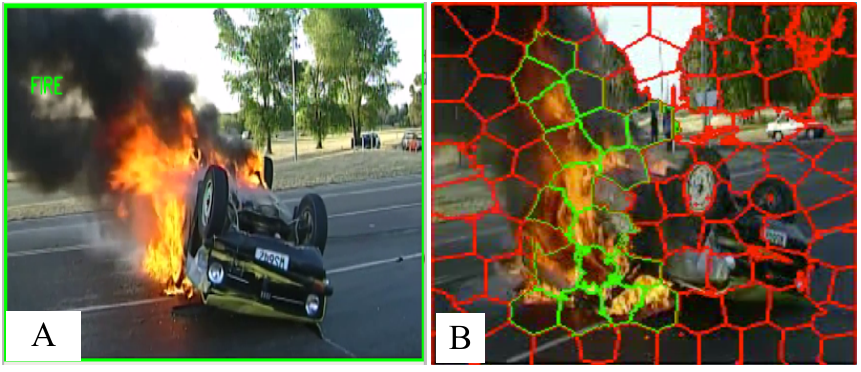}
    \vspace{-0.8cm}
    \caption{An illustrative example of binary full-frame fire detection (A) and localisation via superpixel segmentation (B) (fire = green, no-fire = red).}
    \label{fig:fire_ex}
\end{figure}

Before the advent of deep learning, fire detection work mainly used shallow machine learning approaches using extracted attributes as the input \cite{zhang09fire,Ko2009fire,Chen11:Fire}. The work by \cite{zhang09fire} uses a shallow neural network incorporating a single hidden layer. The model feeds six characteristics of the flame-coloured regions as an input mapping to a hidden layer of seven neurons. A two-class Support Vector Machine (SVM) is used in the work \cite{Ko2009fire} to find a separation between pixels in candidate fire regions to try and remove noise from smoke and differences between frames. Following from this, the work of \cite{Chen11:Fire} develops a non-temporal approach that uses a decision tree as the classification architecture, feeding colour-texture feature descriptors as the input, and achieves 80-90\% mean true positive detection, 7-8\% false positive.

With the advent of deep learning, the focus has shifted from identifying explicit image attributes. A large increase in the classification accuracy has come from creating an end-to-end solution for fire classification \cite{namozov2018efficient,sharma2017deep} and localisation on full-frame/superpixel images (Figure \ref{fig:fire_ex}), achieved by feeding raw image pixel data as input to a Convolutional Neural Network (CNN) architecture. The work of \cite{namozov2018efficient} proposes a custom architecture consisting of convolution, fully connected and pooling layers on a custom dataset using Generative Adversarial Networks (GAN). The work of \cite{sharma2017deep} considers deep CNN architectures such as VGG16 \cite{simonyan2014vgg} and ResNet50 \cite{he2016deep} for the fire detection task. However, these strategies consists of a large number of parameters which lead to slower processing time and may not be suitable for deployment on low-powered embedded devices as commonly found in deployed detection and wide area surveillance systems.

Several CNN architectures \cite{DBLP:journals/corr/IandolaMAHDK16,DBLP:journals/corr/HowardZCKWWAA17,DBLP:journals/corr/abs-1807-11164,DBLP:journals/corr/ZophVSL17} are designed to have a low complexity without compromising on the accuracy. The introduction of depth-wise separable convolutions involves splitting up a conventional $H \times W$ filter into a $H \times 1$ depth-wise filter followed by a $1 \times W$ point-wise filter, vastly reducing the number of floating-point operations and reducing the overall parameter size in a CNN architecture. Recent advancement in creating compact CNN architectures, which focuses on reducing the overhead costs of computation and parameter storage, involves pruning CNN architectures and compressing the weights of various layers \cite{DBLP:journals/corr/LiKDSG16,DBLP:journals/corr/MolchanovTKAK16} without significantly compromising original accuracy. 
In the work of \cite{DBLP:journals/corr/MolchanovTKAK16}, a greedy criteria-based pruning of convolutional kernels by backpropagation is proposed. This strategy \cite{DBLP:journals/corr/MolchanovTKAK16} is computationally efficient and maintains good generalisation in the pruned CNN architecture. 
An approach by \cite{DBLP:journals/corr/LiKDSG16} presents acceleration method for CNN architectures, by pruning convolutional filters that are identified as having a small effect on the output accuracy. By removing entire filters in the network with their connecting feature maps, it prevents an increase in sparsity and reduces computational costs. A one-shot pruning and retraining strategy is adopted in this work \cite{DBLP:journals/corr/LiKDSG16} to save retraining time for pruning filters across multiple layers, which is critical for creating a reduce complexity CNN architecture.

Recent works on non-temporal fire detection \cite{dunnings18fire, samarth19fire} outperform the conventional architectures by simplifying a complex high performance, generalised architecture. The FireNet and InceptionV1Onfire are proposed in the work of \cite{dunnings18fire}, where the architectures are simplified version of AlexNet \cite{NIPS2012_4824} and InceptionV1 \cite{DBLP:journals/corr/SzegedyLJSRAEVR14} respectively. Both architectures offer better performance in fire detection over their parent architectures where FireNet \cite{dunnings18fire} achieves $17$ frames per second (fps) with Accuracy of $0.92$ for the binary classification task. Further architectural advancements, InceptionV3 \cite{szegedy2016rethinking} and InceptionV4 \cite{szegedy2017inception} architectures are experimentally simplified in the most recent work of \cite{samarth19fire}, which achieves Accuracy of $0.96$ for full-frame and $0.94$ for superpixel classification task. Both architectures achieve high accuracy while maintaining a high computational efficiency and throughput.

In this work, we explicitly consider non-temporal fire detection strategy by proposing significantly reduced complexity CNN architectures compared to prior work of \cite{Chen11:Fire,dunnings18fire,samarth19fire}. Our key contributions are the following:
\begin{itemize}
   \item[--] We propose two simplified compact CNN architectures (NasNet-A-OnFire and ShuffleNetV2-OnFire), which are experimentally defined as architectural subsets of seminal CNN architectures \cite{DBLP:journals/corr/ZophVSL17,DBLP:journals/corr/abs-1807-11164} offering maximal performance for the fire detection task.
  \item[--] We employ the proposed compact CNN architectures for (a) binary fire detection, \{{\it fire, no-fire}\}, in full-frame imagery and (b) in-frame classification and localisation of fire using superpixel segmentation \cite{achanta2012slic}.
\end{itemize}
    \section{Proposed Approach}  \label{sec:proposal}
We consider two CNN architectures, NasNet-A-Mobile \cite{DBLP:journals/corr/ZophVSL17} and ShuffleNetV2 \cite{DBLP:journals/corr/abs-1807-11164} (Section \ref{ssec:ref_arch}), which are experimentally optimised using filter pruning \cite{DBLP:journals/corr/LiKDSG16} for fire detection (Section \ref{ssec:sim_arch}). Subsequently, we expand this work for in-frame fire localisation via superpixel segmentation (Section \ref{ssec:superpixel}).    

\subsection{Reference Architectures} \label{ssec:ref_arch}
We select NasNet-A-Mobile \cite{DBLP:journals/corr/ZophVSL17} and ShuffleNetV2 \cite{DBLP:journals/corr/abs-1807-11164} due to their compactness and high performance on ImageNet \cite{deng2009imagenet} classification. Both architectures have high level structures, containing normal cell and reduction cell, however with fundamental differences in how these cells are structured at a low level. Due to the modular structures of the architectures, it is easy to remove/modify different cells.
\begin{figure}[htb!]
    \centering
    \includegraphics[width=8cm]{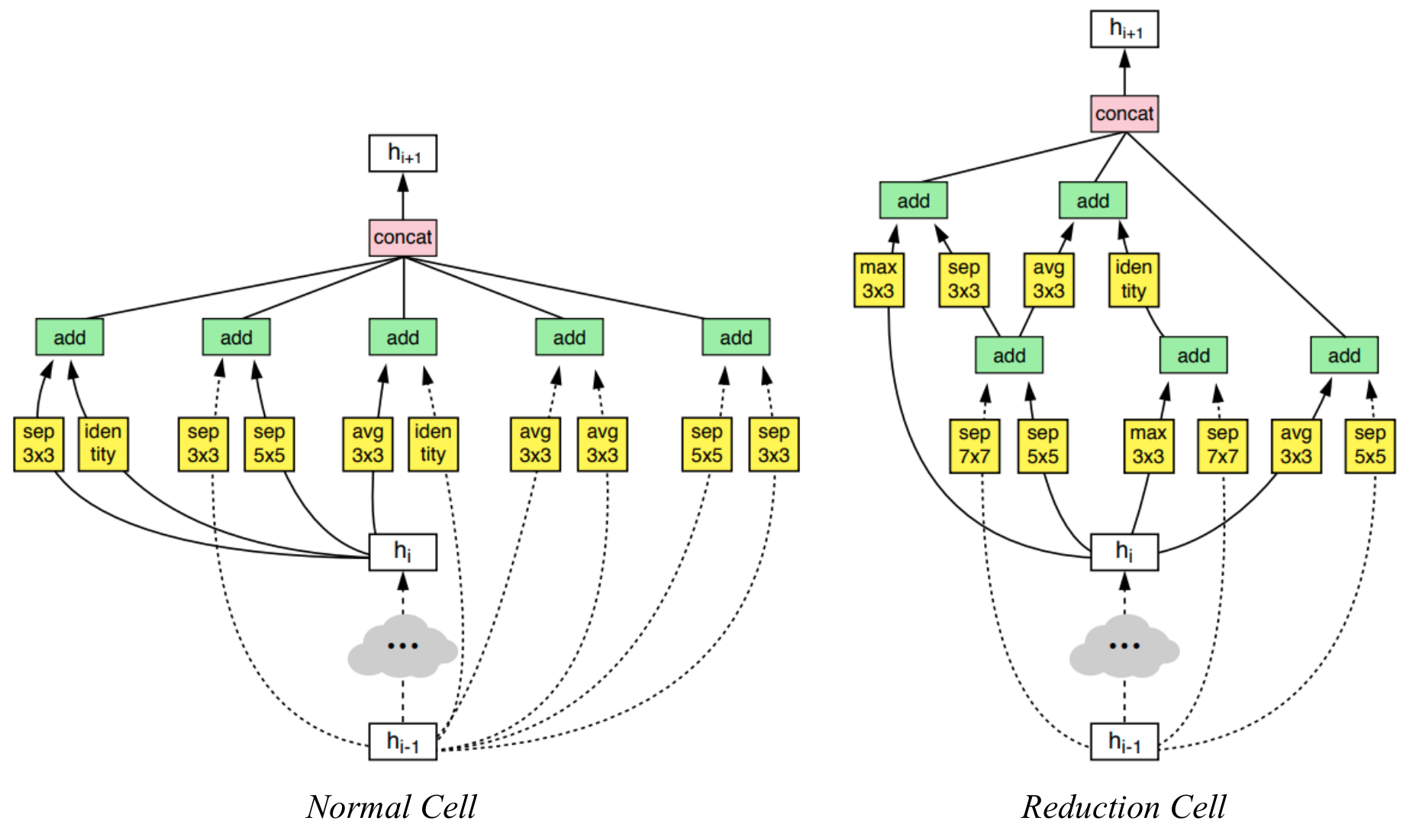}
    \vspace{-0.4cm}
    \caption{Normal and reduction cells of NasNet-A-Mobile \cite{DBLP:journals/corr/ZophVSL17}.}
    \label{fig:nasneta_cells}
\end{figure}

\noindent \textbf{NasNet-A-Mobile} \cite{DBLP:journals/corr/ZophVSL17} consists of an initial $3 \times 3$ convolution layer followed by a sequence repeating three times that consists of a number of reduction cells and four normal cells. 
The normal and reduction cells (Figure \ref{fig:nasneta_cells}), both feed in the input from the previous cell and the cell before, create a residual network. The only convolution layers present in the normal cell are three $3 \times 3$ and two $5 \times 5$ depth-wise separable convolutions. The reduction cell contains one $3 \times 3$, two $5 \times 5$, and two $7 \times 7$ depth-wise separable convolutions. The rest of the layers are either averaging or max pooling layers. 

\begin{figure}[htb!]
    \centering
    \includegraphics[width=6.5cm]{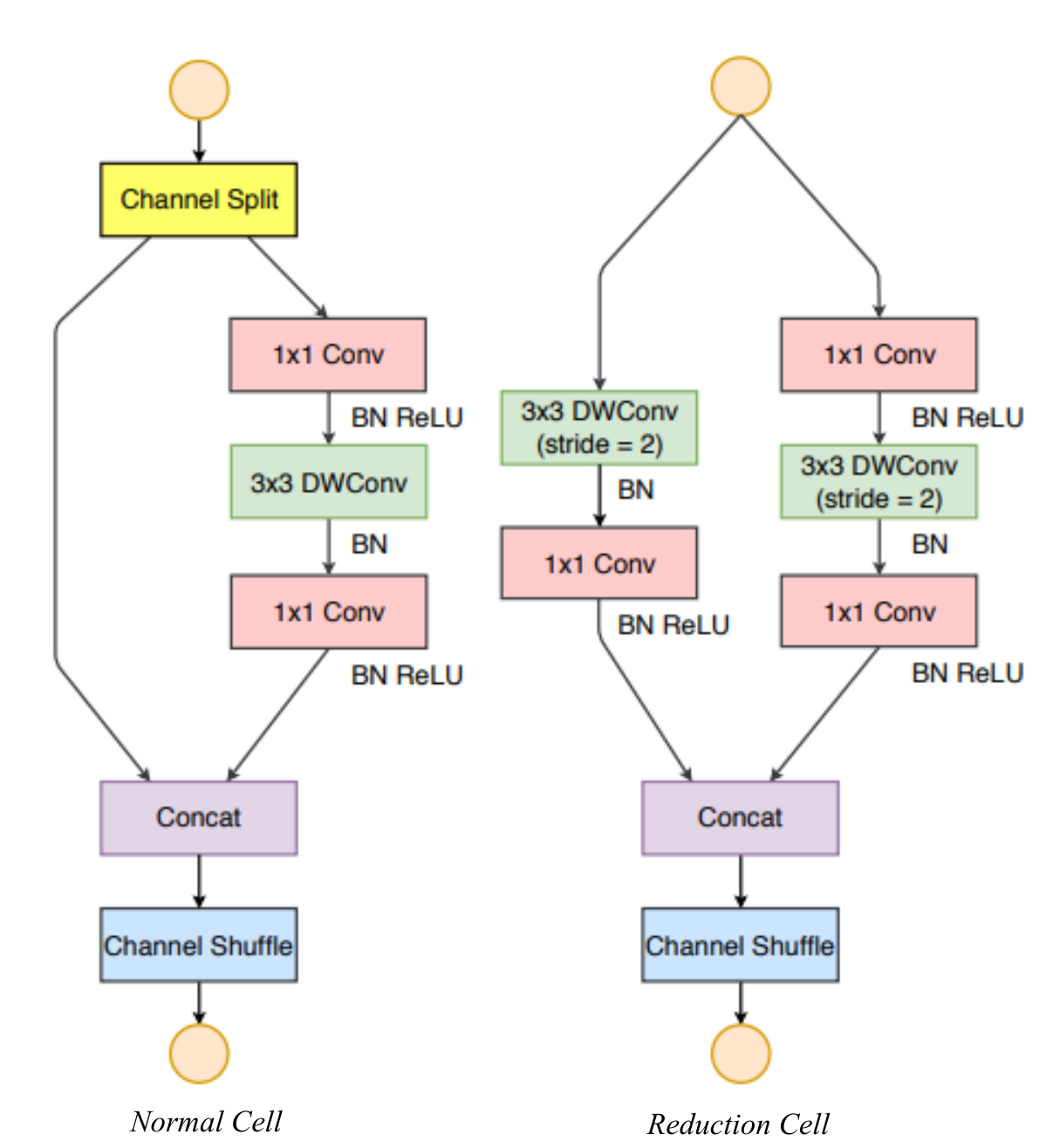}
    \vspace{-0.4cm}
    \caption{Normal and reduction cell of ShuffleNetV2 \cite{DBLP:journals/corr/abs-1807-11164}.}
    \label{fig:shuffle}
\end{figure}

\noindent \textbf{ShuffleNetV2} \cite{DBLP:journals/corr/abs-1807-11164} consists of an initial $3 \times 3$ convolution layer followed by a $3 \times 3$ max pooling layer. This is followed by three reduction and normal cells (Figure \ref{fig:shuffle}). There is only one reduction cell for each loop and the number of normal cells is $[3, 7, 3]$. This is followed by a final point-wise convolution and a global pooling layer. The normal cell starts by splitting the number of channels in half, where one half is unchanged and the other half has three convolutions with two point-wise $1 \times 1$ convolutions and a $3 \times 3$ depth-wise convolution. The output dimension is equal to the input in the normal cell. The channels are concatenated and shuffled in order to mix the channels across the branches. This is applied in both halves of the cells. 
The reduction cell does not split the channels and both branches receive the whole input. The right branch of the reduction cell is similar to the right branch of the normal cell however the depth-wise convolution has a stride of $2$ to reduce the height and width dimensions by two. The use of point-wise convolutions and depth-wise convolutions allow the network to go very deep without the number of parameters blowing up. 
\vspace{-0.2cm}
\subsection{Simplified CNN Architectures} \label{ssec:sim_arch}
We experimentally investigate variations in architectural configurations of each reference architecture (Section \ref{ssec:ref_arch}).

In the simplified CNN architectures, we use transfer learning from network models training on ImageNet \cite{deng2009imagenet} by removing the final fully connected layer from both architectures \cite{DBLP:journals/corr/ZophVSL17,DBLP:journals/corr/abs-1807-11164} and create a new linear layer that mapped to a single value for binary \{{\it fire, no-fire}\} classification. The entire architecture is frozen except the final linear layer for the first half of training. Subsequently, we unfreeze the final convolution layer for ShuffleNetV2 \cite{DBLP:journals/corr/abs-1807-11164}, and the final two reductions, normal cell iterations for NasNet-A-Mobile \cite{DBLP:journals/corr/ZophVSL17}. This prevents the network over-fitting during training and allowing us to train the model longer for better generalisation.  

\subsubsection{Simplified NasNet-A-Mobile}  \label{sssec:nas}
\begin{figure*}[htb!]
  \centering
  \includegraphics[width=15cm]{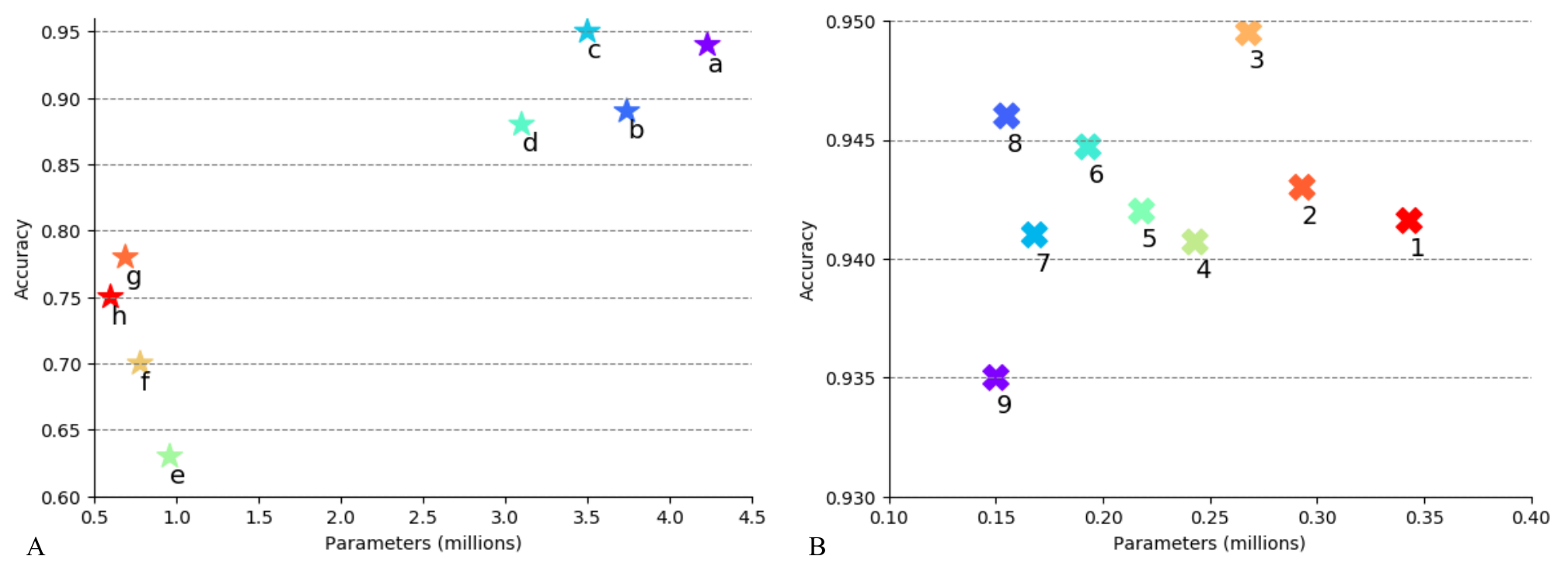}
  \vspace{-0.4cm}
  \caption{Parameter size and accuracy comparison for the two architecture variants:  (A) Nasnet-A-Mobile where a to e represents $NasNet_{v01}$ - $NasNet_{v08}$. (B) ShuffleNetV2 where 1-9 represents $ShuffleNetV2_{v01}$ - $ShuffleNetV2_{v09}$.}
  \label{fig:graphs}    
\end{figure*}

The base model for NasNet-A-Mobile \cite{DBLP:journals/corr/ZophVSL17} is pre-trained with $1,056$ output channels for ImageNet \cite{deng2009imagenet} classification. The main experimentation of this architecture revolves around the number of normal cells in the model. 
\begin{table}[htb]
\centering
\renewcommand*{\arraystretch}{0.85}
\caption{ NasNet-A-Mobile variants with different components.}
\vspace{-0.3cm}
\begin{tabular}{lccccc}
\hline
{\scriptsize Architecture} & {\scriptsize $A_{N}=2$} & {\scriptsize $A_{N}=4$} & {\scriptsize $A_{3} = A_{N}$} & {\scriptsize $A_{3}=0$} & {\scriptsize {\shortstack[l]{Reduced \\ Filter}}} \\ \hline
\scriptsize $NasNet_{v01}$ & & {\color{green} \checkmark } & {\color{green} \checkmark } & & \\
\scriptsize $NasNet_{v02}$ & & {\color{green} \checkmark } & & {\color{green} \checkmark } & \\

\scriptsize{$\underline{NasNet_{v03}}$} & {\color{green} \checkmark } &  & {\color{green} \checkmark } & & \\
\scriptsize $NasNet_{v04}$ & {\color{green} \checkmark } & & & {\color{green} \checkmark } & \\
\scriptsize $NasNet_{v05}$ & & {\color{green} \checkmark } & {\color{green} \checkmark } & & {\color{green} \checkmark } \\
\scriptsize $NasNet_{v06}$ & & {\color{green} \checkmark } & & {\color{green} \checkmark } & {\color{green} \checkmark } \\
\scriptsize $NasNet_{v07}$ & {\color{green}
\checkmark } &  & {\color{green} \checkmark } & & {\color{green} \checkmark } \\
\scriptsize $NasNet_{v08}$ & {\color{green} \checkmark } & & & {\color{green} \checkmark } & {\color{green} \checkmark } \\ \hline
\end{tabular}
\label{tab:nasnet_variants}
\end{table}
In our simplified NasNet-A-Mobile architectures, we experiment with eight different variants, with four different architectural structure differences with and without reduced filters (Table \ref{tab:nasnet_variants}). The number of filters is calculated by the number of penultimate filters specified in the architecture. We reduce this number from $1,056$ to $480$ penultimate filters for the reduced filter variants. This creates a reduction of $60\%$ of the filters throughout the whole model. This drastically reduces the number of parameters but achieves a very low accuracy for the each of the reduced filter variants (Figure \ref{fig:graphs}-A). 
There is a sharp difference in accuracy between the reduced filter variants and full filter variants in the NasNet-A-Mobile architecture \cite{DBLP:journals/corr/ZophVSL17} (points \{a,b,c,d\} vs. points \{e,f,g,h\} in Figure \ref{fig:graphs}-A) with the highest reduce filter variant achieving $0.77$ accuracy compared to $0.95$ for the corresponding full filter variant.

\subsubsection{Simplified ShuffleNetV2} \label{sssec:shuf}
The number parameters in  ShuffleNetV2 architecture \cite{DBLP:journals/corr/abs-1807-11164} are $340,000$. Upon examining the distribution of parameters in the architecture, over $200,000$ of the parameters are contained in the final convolutional layer. 
Therefore we freeze the parameters in the first half of the network and experimentally incorporate the filter pruning strategy to further reduce the complexity of the final convolutional layer, without compromising the accuracy. We adopt a similar approach proposed in the work of \cite{DBLP:journals/corr/LiKDSG16}, which computes the L2-normalisation of the filters, and subsequently we sort and remove the lower valued filters. The intuition behind this strategy is that filters with lower values will be less effective to the final output of the architecture. The model is further retrained with the removed pruned filters.       
\begin{table}[htb!]
\centering
\renewcommand*{\arraystretch}{0.85}
\caption{ShuffleNetV2 \protect\cite{DBLP:journals/corr/abs-1807-11164} pruning variants on the final convolutional filters.}
\vspace{-0.3cm}
\begin{tabular}{lccc}
\hline
\small Architecture & \shortstack[l]{\small Pruned \\ \small Filters} & \shortstack[l]{\small Final Convolutional \\ \small Layer Parameters} & \shortstack[l]{\small Total \\ \small Parameters} \\ \hline
\scriptsize $ShuffleNet_{v01}$  & \small 128 & \small 196,608 & \small 342,897 \\
\scriptsize $ShuffleNet_{v02}$ & \small 256 & \small 147,456  & \small 292,897 \\
\scriptsize $ShuffleNet_{v03}$ & \small 384 & \small 122,800 & \small 267,937 \\
\scriptsize $ShuffleNet_{v04}$ & \small 512  & \small 98,304 & \small 242,977 \\
\scriptsize $ShuffleNet_{v05}$ & \small 640 & \small 73,728 & \small 218,017 \\
\scriptsize $ShuffleNet_{v06}$ & \small 768 & \small 49,152 & \small 193,057 \\
\scriptsize $ShuffleNet_{v07}$ & \small 896 & \small 24,576 & \small 168,097 \\
\scriptsize $\underline{ShuffleNet_{v08}}$ & \small 960 & \small 12,288 &  \small 155,617 \\ 
\scriptsize $ShuffleNet_{v09}$ & \small 992 & \small 6,144 & \small 149,377 \\ \hline
\end{tabular}
\label{tab:shufflenet_reduction}
\end{table}

Table \ref{tab:shufflenet_reduction} shows the number of filters pruned in the different variants of the ShuffleNetV2 architecture. We start by pruning $128$ filters that represents $1/8th$ of the number of filters in the final convolutional layer. We remove $128$ filters in each iteration and continued as long as the accuracy does not degrade ($ShuffleNet_{v01}$ to $ShuffleNet_{v07}$). We subsequently prune a further $64$ filters in $ShuffleNet_{v08}$ and $32$ filters in $ShuffleNet_{v09}$ variants however at this stage we stop  pruning due to a decrease in accuracy (points 8,9 in Figure \ref{fig:graphs}-B).

\begin{figure}[htb]
    \centering
    \includegraphics[width=\linewidth]{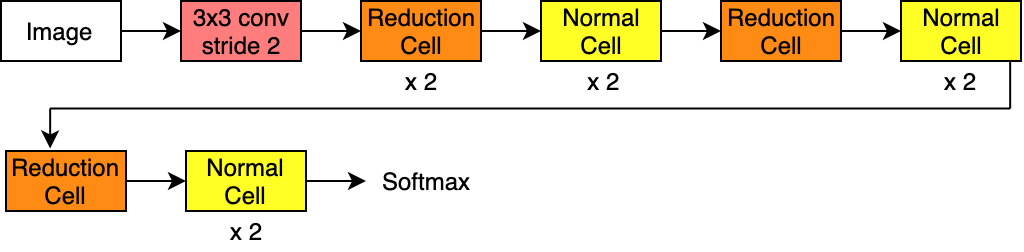}
    \vspace{-0.8cm}
    \caption{Reduced complexity architecture for NasNet-A-OnFire optimised for fire detection.}
    \label{fig:nasnetarchitecture_reduced}
\end{figure}

With exhaustive experimentation using both architectures, we propose following two reduced complexity architectural variants modified for the binary \{{\it fire, no-fire}\} classification task.

\noindent \textbf{NasNet-A-OnFire} is a variant based on $NasNet_{v03}$ such that each group of normal cells in the model only contain two normal cells compared to the previous four (Table \ref{tab:nasnet_variants}, highlighted in underline). The normal cells and reduction cells in the NasNet-A-OnFire architecture remain the same as shown in Figure \ref{fig:nasneta_cells}. The total number of parameter in NasNet-A-OnFire is $3.2$ million. 

\begin{figure}[htb]
    \centering
    \includegraphics[width=\linewidth]{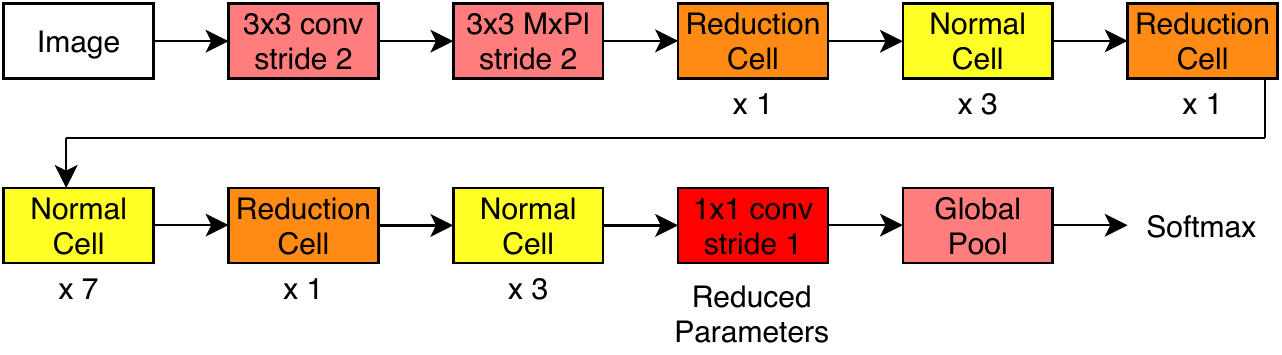}
    \vspace{-0.8cm}
    \caption{Reduced complexity architecture for ShuffleNetV2-OnFire optimised for fire detection.}
    
    \label{fig:ShuffleNetV2-OnFire}
\end{figure}

\noindent \textbf{ShuffleNetV2-OnFire} is a variant of $ShuffleNet_{v08}$ (Table \ref{tab:shufflenet_reduction}, highlighted in underline) with the same fundamental architecture design as ShuffleNetV2. In ShuffleNetV2-OnFire (Figure \ref{fig:ShuffleNetV2-OnFire}), by applying filter pruning strategy, we reduce the number of filters in the final convolutional layer, which leads to a total number of the model parameter of only $155,617$.

We employ these two CNN architectures for \{{\it fire, no-fire}\} classification task applied on full frame binary and superpixel segmentation based fire detection.

\vspace{-0.0cm}
\subsection{Superpixel Localisation} \label{ssec:superpixel}
We further expand this work fo in-image fire localisation by using superpixel regions \cite{achanta2012slic}, contrary to the earlier work \cite{Phillips02fire,CELIK2009fire} that largely relies on colour based initial locatiosation. Superpixel based techniques over-segment an image into perceptually meaningful regions which are similar in colour and texture. We use Simple Linear Iterative Clustering (SLIC) \cite{achanta2012slic}, which performs iterative clustering in a similar manner to {\it k-means} to reduced spatial dimensions, where the image is segmented into approximately equally sized superpixels (Figure \ref{fig:fire_ex}-B). Each superpixel region is classified using proposed NasNet-A-OnFire/ShuffleNetV2-OnFire architecture formulated as a \{{\it fire, no-fire}\}, for fire detection task.
    \section{Evaluation} \label{sec:evaluation}
\subsection{Experimental Setup} \label{sec:exp}
In this section, we present the dataset and implementation details used in our experiments. 
\subsubsection{Dataset} \label{ssec:db}
We use the dataset created in the work \cite{dunnings18fire} to train and evaluate the performance of our proposed architectures.
The dataset consists of $26,339$ full-frame images (Figure \ref{fig:fire_ex}-A) of size $224 \times 224$, with $14,266$ images of fire, and $12,073$ images of non-fire. Training is performed over a set of $23,408$ images (70 : 20 data  split) and testing reported over $2,931$ images. The superpixel (SLIC \cite{achanta2012slic}) training set consists of $54,856$ fire, and $167,400$ non-fire superpixels with a test set of $1178$ fire and $881$ non-fire examples.

\subsubsection{Implementation Details} \label{ssec:implenmentation}
The proposed architectures are implemented in PyTorch \cite{paszke2019pytorch} and trained with the following configuration: backpropagation optimisation performed via Stochastic Gradient Descent (SGD), binary cross entropy loss function, learning rate (lr) of $0.0005$, and $40$ epochs. We measure the performance using the following CPU and GPU configuration: Intel Core i5 with 8GB of RAM CPU, and NVIDIA 2080Ti GPU.

\subsection{Results} \label{ssec:results}
We present the results of the simplified architectures compared to the state-of-the-art for binary classification (Section \ref{ssec:bin_cls}) and superpixel localisation task (Section \ref{ssec:super_loc}). For statistically comparing different architectures, we use the metrics of true positive rate (TPR), false positive rate (FPR), F-score (F), Precision (P) and Accuracy (A), Complexity (number of parameters in millions, C), the ratio between accuracy and the number of parameters in the architecture (A:C) and achievable frames per second (fps) throughput.

\subsubsection{Binary Classification} \label{ssec:bin_cls}
From the results presented in Table \ref{tab:bin_classification} for full-frame \{{\it fire, no-fire}\} classification task, our proposed architectures achieve comparable performance with prior works \cite{dunnings18fire,samarth19fire}. We present only the highest performing variants, NasNet-A-OnFire and ShuffleNetV2-OnFire, in our experimentation (Table \ref{tab:bin_classification}-lower).

\begin{table}[htb!]
\centering
\renewcommand*{\arraystretch}{0.85}
\caption{The statistical performance for full-frame binary fire detection. Upper: Prior works. Lower: Our approaches.}
\vspace{-0.3cm}
\begin{tabular}{llllll}
\hline 
\small Architecture & \small TPR  & \small FPR  & \small F   & \small P  & \small A    \\ \hline 
\rowcolor{blue!4}
\scriptsize{FireNet \cite{dunnings18fire}} & \small 0.92 & \small 0.09 & \small 0.93 & \small 0.93 & \small 0.92 \\ 
\rowcolor{blue!8}
\scriptsize{Inception V1-OnFire \cite{dunnings18fire}} & \small \textbf{0.96} & \small 0.10 & \small 0.94 & \small 0.93 & \small 0.93 \\
\rowcolor{blue!4}
\scriptsize{Inception V3-OnFire \cite{samarth19fire}} & \small 0.95 & \small 0.07 & \small 0.95 & \small 0.95 & \small 0.94 \\ 
\rowcolor{blue!8}
\scriptsize{Inception V4-OnFire \cite{samarth19fire}} & \small 0.95 & \small 0.04 & \small \textbf{0.96} & \small \textbf{0.97} & \small \textbf{0.96} \\ \hline
\rowcolor{blue!14}
\scriptsize{\textbf{NasNet-A-OnFire}} & \small 0.92 & \small \textbf{0.03} & \small 0.94 & \small 0.96 & \small 0.95 \\ 
\rowcolor{blue!14}
\scriptsize{\textbf{ShuffleNetV2-OnFire}} & \small 0.93 & \small 0.05 & \small 0.94 & \small 0.94 & \small 0.95 \\ \hline
\end{tabular}
\label{tab:bin_classification}
\end{table}

Both the architectures achieve a FPR less than or equal to the minimum FPR in the current state-of-the-art approaches \cite{dunnings18fire,samarth19fire} (Table \ref{tab:bin_classification}-upper) with ShuffleNetV2-OnFire achieving FPR: $0.05$ and NasNet-A-OnFire with FPR: $0.03$ (compared to Inception V4-OnFire \cite{samarth19fire} with FPR: $0.04$). The overall accuracy remains consistent with both proposed architectures achieving A: $0.95$. Although there is a slight decrease in TPR for both architectures (Table \ref{tab:bin_classification}-lower), still it achieves comparable TPR with prior works considering the compactness and reduced complexity of the architectures as presented in Tables \ref{tab:bin_cllas} and \ref{tab:super_pix_speed}. NasNet-A-OnFire offers the best performance in terms of FPR (FPR: 0.03), however ShuffleNetV2-OnFire achieves better TPR with $0.93$. ShuffleNetV2-OnFire outperforms InceptionV3-OnFire \cite{samarth19fire} at FPR (FPR: 0.05 vs 0.07) and accuracy (A: 0.95 vs 0.94), both being the smallest architectures (Table \ref{tab:super_pix_speed}). 

\subsubsection{Superpixel Localisation} \label{ssec:super_loc}
For superpixel based fire detection, NasNet-A-OnFire achieves the highest TPR with $0.98$ (Table \ref{tab:superpixel_classification}-lower) outperforming the prior work achieving TPR: $0.94$ (InceptionV3-OnFire/InceptionV4-OnFire \cite{samarth19fire}, Table \ref{tab:superpixel_classification}-upper). However NasNet-A-OnFire suffers from a high FPR of $0.15$ compared to InceptionV3-OnFire/InceptionV4-OnFire \cite{samarth19fire} with FPR: 0.07 and FPR: 0.06 respectively. ShuffleNetV2-OnFire achieves a lower FPR (FPR: 0.08, Table \ref{tab:superpixel_classification}-lower), which is comparable to prior work \cite{dunnings18fire,samarth19fire} and achieves a similar accuracy (A: 0.94). 
Both of our architectures achieve a higher F-score (F: $0.98$) and Precision (P: $0.99$) outperforming prior work \cite{dunnings18fire,samarth19fire}.

\begin{table}[htb!]
\centering
\renewcommand*{\arraystretch}{0.85}
\caption{Localisation results using within frame superpixel approach. Upper: Prior works. Lower: Our approaches.}
\vspace{-0.3cm}
\begin{tabular}{llllll}
\hline
\small Architecture & \small  TPR  & \small  FPR  & \small  F   & \small  P    & \small  A    \\ \hline 
\rowcolor{blue!4}
\scriptsize{InceptionV1-OnFire \cite{dunnings18fire}} & \small  0.92 & \small  0.17 & \small  0.9 & \small  0.88 & \small  0.89 \\ 
\rowcolor{blue!8}
\scriptsize{InceptionV3-OnFire \cite{samarth19fire}} & \small  0.94 & \small  0.07 & \small  0.94 & \small  0.93 & \small  0.94 \\ 
\rowcolor{blue!4}
\scriptsize{InceptionV4-OnFire \cite{samarth19fire}} & \small  0.94 & \small  \textbf{0.06} & \small  0.94 & \small  0.94 & \small  0.94 \\ \hline
\rowcolor{blue!14}
\scriptsize{\textbf{NasNet-A-OnFire}} & \small  \textbf{0.98} & \small  0.15 & \small  \textbf{0.98} & \small  \textbf{0.99} & \small  \textbf{0.97} \\ 
\rowcolor{blue!14}
\scriptsize{\textbf{ShuffleNetV2-OnFire}} & \small  0.94 & \small  0.08 & \small  0.97 & \small  \textbf{0.99} & \small  0.94 \\ \hline
\end{tabular}
\label{tab:superpixel_classification}
\end{table}

\begin{figure}[htb!]
        \centering
    \includegraphics[width=\linewidth]{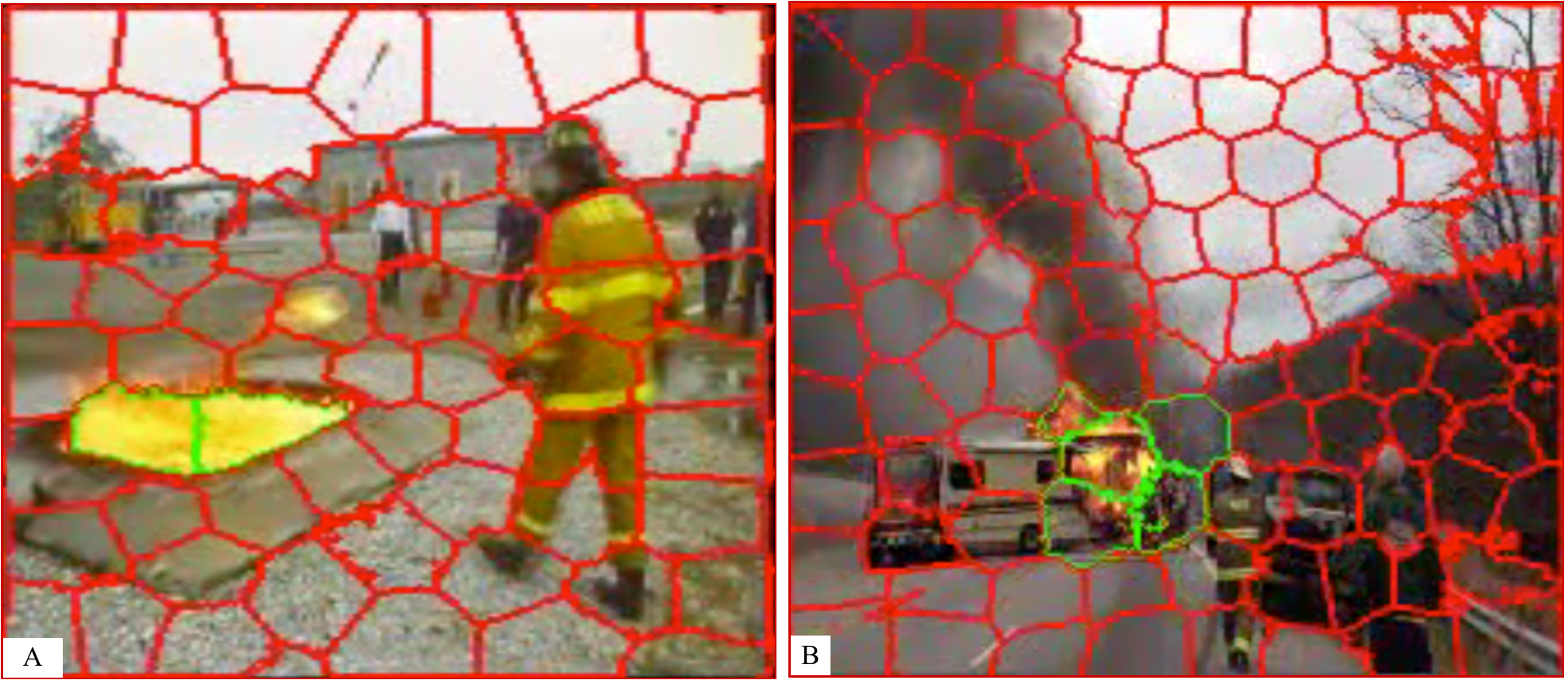}
    \vspace{-0.8cm}
    \caption{Exemplar superpixel based fire localisation using ShuffleNetV2-OnFire on two challenging scenarios: (A) image containing a fireman wearing an outfit similar colour to the fire and (B) image containing a red colour truck (fire = green, no-fire = red).}
    \label{fig:sp_tp}
\end{figure}

Qualitative examples of fire localisation, using ShuffleNetV2-OnFire, are illustrated in Figure \ref{fig:sp_tp}. Each example presents challenging scenarios that could lead to false positive detection. These examples demonstrate the robustness of the proposed architecture for the fire detection task in various challenging scenarios.

\subsubsection{Architecture Simplification vs Speed} \label{ssec:model_speed}
Table \ref{tab:bin_cllas} presents the computational efficiency and speed for full-frame classification using different architectures. ShuffleNetV2-OnFire improves on the computational efficiency by $7.8$ times achieving $608.97$ compared to InceptionV1-OnFire \cite{dunnings18fire} achieving $77.9$ while running on CPU configuration. It also improves on classification speed (fps:  $40$) compared to FireNet \cite{dunnings18fire} (fps: $17$), while having $437\times$ fewer number of parameters (C: $0.156$ million), and achieving a higher accuracy (A: $95$\%). Whilst computing on a GPU configuration, the inference time increases further for ShuffleNetV2-OnFire achieving $69$ fps, and NasNet-A-OnFire achieves $35$ fps (Table \ref{tab:bin_cllas}-lower). Overall ShuffleNetV2-OnFire is the best performing architecture for full-frame binary classification in terms of accuracy and efficiency (A:C: $608.7$), outperforming prior works \cite{dunnings18fire}.

\begin{table}[htb!]
\centering
\renewcommand*{\arraystretch}{0.85}
\caption{Computational efficiency for full-frame classification.}
\vspace{-0.3cm}
\begin{tabular}{lllll}
\hline
\small Architecture       & \small  C    & \small  A(\%) & \small  A:C    & \small  fps  \\ \hline 
\rowcolor{blue!4}
\scriptsize{FireNet \cite{dunnings18fire}}  & \small  68.3 & \small  91.5  & \small  1.3    & \small  17   \\
\rowcolor{blue!8}
\scriptsize{InceptionV1-OnFire \cite{dunnings18fire}} & \small  1.2  & \small  93.4  & \small  77.9   & \small  8.4  \\ \hline
\rowcolor{blue!14}
\scriptsize{\textbf{NasNet-A-OnFire}}  & \small  3.2  & \small  \textbf{95.3}  & \small  29.78  & \small  7 \\
\rowcolor{blue!14}
\scriptsize{\textbf{ShuffleNetV2-OnFire}}  & \small  \textbf{0.156} & \small  95   & \small  \textbf{608.97} & \small  \textbf{40}   \\ \hline
\rowcolor{blue!14}
\scriptsize{\textbf{NasNet-A-Mobile (GPU)}}         & \small  3.2  & \small  \textbf{95.3}  & \small  29.78  & \small  35 \\ 
\rowcolor{blue!14}
\scriptsize{\textbf{ShuffleNetV2-OnFire (GPU)}}     & \small  \textbf{0.156} & \small  95   & \small  \textbf{608.97} & \small  \textbf{69}   \\ \hline
\end{tabular}
\label{tab:bin_cllas}
\vspace{0.05cm}
\caption{Computational efficiency for superpixel localisation.}
\vspace{-0.3cm}
\begin{tabular}{lllll}
\hline
\small Architecture & \small  C & \small  A(\%) & \small  A:C    & \small  fps  \\ \hline 
\rowcolor{blue!4}
\scriptsize{InceptionV3-OnFire \cite{samarth19fire}} & \small  0.96 & \small  94.4  & \small  98.09  & \small  13.8 \\
\rowcolor{blue!8}
\scriptsize{InceptionV4-OnFire \cite{samarth19fire}} & \small  7.18 & \small  95.6  & \small  13.37  & \small  12   \\ \hline
\rowcolor{blue!14}
\scriptsize{\textbf{NasNet-A-OnFire (GPU)}}    & \small  3.2  & \small  \textbf{97.1}  & \small  30.34  & \small  5      \\ 
\rowcolor{blue!14}
\scriptsize{\textbf{ShuffleNetV2-OnFire (GPU)}} & \small  \textbf{0.156} & \small  94.4   & \small  \textbf{605.13} & \small  \textbf{18} \\ \hline
\end{tabular}
\label{tab:super_pix_speed}
\end{table}

In Table \ref{tab:super_pix_speed}, we present the computational efficiency of superpixel localisation (all superpixels are processed for each frame) running  on a GPU configuration. Although NasNet-A-OnFire achieves the highest accuracy obtaining A: $97.1$, however it operates at the lowest fps (fps: $5$). ShuffleNetV2-OnFire, consists of only $0.156$ million parameter, provides the maximal throughput of $18$ fps, outperforming InceptionV3-OnFire \cite{samarth19fire} (fps: $13.8$), as presented in Table \ref{tab:super_pix_speed}-lower.    
We also measure the runtime of our proposed architectures on low-powered embedded device (Nvidia Xavier-NX \cite{xavier_nx}) as presented in Table \ref{tab:xaviernx_speed}. The PyTorch implementation operates at a lower speed (fps: 6 \& 18) compared to standard CPU/GPU implementation. However, conversion to 16-bit floating point numerical accuracy (via TensorRT, FP16) improves the inference time by $\sim$3-6 times, achieving fps of $35$ (NasNet-A-OnFire) and $49$ (ShuffleNetV2-OnFire), without compromising the performance accuracy. 
\begin{table}[htb!]
\centering
\renewcommand*{\arraystretch}{0.8}
\caption{Runtime (fps) comparison of \{{\it fire, no-fire}\} classification on different hardware devices.}
\vspace{-0.3cm}
\begin{tabular}{lcccc}
\hline
\multirow{2}{*}{\small Architecture} & \small CPU & \small GPU & \multicolumn{2}{c}{\small Xavier-NX}  \\ \cline{2-5}
& \small PyTorch & \small PyTorch & \small PyTorch & \small TensorRT  \\ \hline 
\rowcolor{blue!4}
\scriptsize NasNet-A-OnFire & \small  7 & \small  35 & \small  6 & \small  \textbf{35} \\ 
\rowcolor{blue!14}
\scriptsize ShuffleNetV2-OnFire & \small  40 & \small  69 & \small  18 & \small  \textbf{49}  \\ \hline
\end{tabular}
\label{tab:xaviernx_speed}
\end{table}
    \section{Conclusion} \label{sec:conclusion}
We propose a compact and reduced complexity CNN architecture (ShuffleNetV2-OnFire) that is over six times more compact than the prior works \cite{dunnings18fire,samarth19fire} for fire detection with a size of $\sim$0.15 million parameters. This significantly outperforms prior works by operating over $\sim$2 times faster. Proposed CNN architectures (NasNet-A-OnFire and ShuffleNetV2-OnFire) are not only compact in size but also achieve similar performance accuracy with $95$\% for full-frame and $94.4$\% for superpixel based fire detection. Subsequently, implementation on low-powered devices (achieving $49$ fps) makes our architectures suitable for various real-world applications. Overall, we illustrate a strategy for simplifying the CNN architectures through experimental analysis and filter pruning while maintaining the accuracy and increasing the computational efficiency. Future work will focus on additional synthetic image data training via Generative Adversarial Networks.

    {
\bibliographystyle{IEEEtran}
\bibliography{refs,icip-2016-refs,projectpaper}
}
\end{document}